\documentclass[letterpaper,10pt,conference]{ieeeconf}

\IEEEoverridecommandlockouts
\usepackage{iftex}
\ifXeTeX
  \usepackage{fontspec}
\fi

\usepackage{amsmath,amssymb}
\usepackage{graphicx}
\usepackage[caption=false,font=footnotesize]{subfig}
\usepackage{booktabs}
\usepackage{multirow}
\usepackage{array}
\usepackage{xcolor}
\usepackage{url}

\newcommand{\method}{UpDown-SC}

\newcommand{\best}[1]{\ensuremath{\mathbf{#1}}}

\title{\LARGE \bf
UpDown-SC: Gravity-Canonicalized Dual-Envelope\\
Scan Context for Indoor LiDAR Place Recognition
}

\author{Jie Xu, Yongxin Yang, Ziyi Jin, Kangjin Yu, Hongjun Huang, Chao Han, and Zhongpu Xia%
\thanks{The authors are with Anyverse Dynamics, Beijing, China
(e-mail: jeff\_xu\_0503@foxmail.com). Corresponding author: Zhongpu Xia.}%
}

\begin{document}
\maketitle
\thispagestyle{empty}
\pagestyle{empty}

%%%%%%%%%%%%%%%%%%%%%%%%%%%%%%%%%%%%%%%%%%%%%%%%%%%%%%%%%%%%%%%%%%%%%%%%%%%%%%%%
\begin{abstract}
LiDAR place recognition is a key front end for loop closure and global relocalization, yet indoor retrieval remains difficult when attitude or sensor mounting height changes between mapping and query sessions. Scan Context stores the maximum height in each polar cell; indoors, broad ceilings can suppress the lower and mid-level geometry that distinguishes adjacent rooms and corridors. We present \method, a training-free polar descriptor that first canonicalizes gravity and then represents two complementary surfaces: the upper envelope of lower/middle structures and the lower envelope of overhead structures. Their physical split is estimated once from a cell-balanced map height distribution and reused by every query. A mask-aware, non-uniform two-channel distance retains discriminative lower-level evidence while limiting sensitivity to its cross-session variation, without treating unobserved cells as zero-height measurements. Conventional Scan Context shortlisting and circular yaw alignment are retained, so retrieved hypotheses directly initialize geometric verification. Experiments across repeated indoor sessions, mounting-height changes, mixed outdoor-to-indoor trajectories, and an outdoor transfer sequence show more reliable first-choice retrieval on the indoor and mounting-height-varied sessions. A paired test finds a significant gain over Scan Context on the in-house sessions. \method\ also gives the best or second-best F1max and AUPR under threshold-based acceptance while retaining a lightweight CPU front end. Continuous replay confirms that the retrieved hypotheses support metric prior-map localization. Code and evaluation artifacts: \url{https://github.com/jiejie567/updown-sc}.
\end{abstract}

%%%%%%%%%%%%%%%%%%%%%%%%%%%%%%%%%%%%%%%%%%%%%%%%%%%%%%%%%%%%%%%%%%%%%%%%%%%%%%%%
\section{Introduction}

LiDAR place recognition (LPR) associates a current observation with a previously mapped place, supplying loop closures or global pose hypotheses to a localization back end. In indoor service robotics, a map is often reused after the sensor is remounted or its motion changes, so queries at the same place can differ in mounting height, body occlusion, roll, or pitch.

Scan Context (SC) provides an efficient and training-free solution by partitioning an egocentric point cloud into polar ring--sector cells and storing the maximum height in each cell~\cite{kim2018scancontext}. Its circular sector search naturally handles heading changes and returns a yaw estimate. The maximum operator that works well in outdoor streets can become less informative indoors: a ceiling may occupy many cells and repeatedly become their maximum, suppressing the shelves, door frames, partitions, and furniture below it. The resulting descriptor is stable but insufficiently distinctive, especially along corridors with similar ceiling geometry. SC also assumes that roll and pitch are mild, an assumption stated explicitly by subsequent SC variants~\cite{kim2022scancontextpp}.

Mounting variation adds a second difficulty: a sensor-frame height split moves relative to the scene when the LiDAR is remounted, and roll and pitch mix vertical and horizontal structure. We therefore use gravity and a measured descriptor-origin height to express vertical structure consistently across acquisitions, and estimate the physical split from the map instead of tuning an environment-specific threshold.

\begin{figure}[t]
    \centering
    \includegraphics[width=\columnwidth]{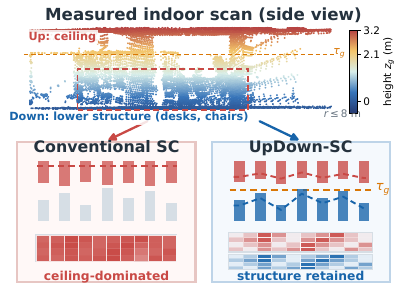}
    \caption{Why maximum-height SC loses indoor structure. Top: a
    gravity-canonicalized side view of one measured 0.1-s indoor query,
    colored by height. The dashed line marks the map split $\tau_g$; the display
    omits far and above-ceiling returns. Where lower
    and overhead structure share a polar cell, SC keeps only the upper
    maximum. \method\ keeps both envelopes. Lower panels are schematics, not
    measured performance.}
    \label{fig:motivation}
\end{figure}

\begin{figure*}[!t]
    \centering
    \includegraphics[width=\textwidth]{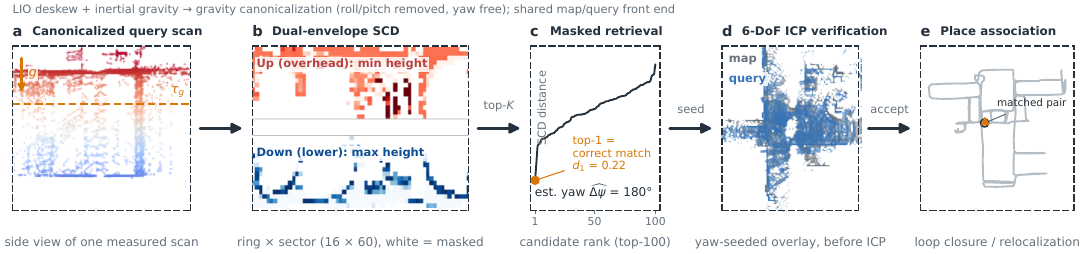}
    \caption{End-to-end pipeline on one measured IH query from the 2~m
    evaluation. The shared map/query front end canonicalizes gravity and
    builds the dual-envelope SCD. Mask-aware retrieval returns ranked place
    and yaw hypotheses (here $\widehat{\Delta\psi}=180^\circ$), and
    point-to-map ICP verifies each 6-DoF seed. Panel (d) overlays the query
    on the retrieved keyframe using only the descriptor seed, before ICP.
    Retrieval metrics are measured before ICP.}
    \label{fig:method_workflow}
\end{figure*}

We address these issues with \method, a gravity-canonicalized dual-envelope extension of SC. Each deskewed frame is restored to the descriptor-origin body frame and leveled while yaw remains free, with the measured descriptor-origin height expressing returns relative to the ground. A constrained, cell-balanced Otsu criterion selects one map-level split that is stored with the database and reused for queries. Each polar cell records the highest return below this split and the lowest return above it, preventing overhead structure from overwriting lower geometry. A masked, fixed non-uniform distance retains the more variable lower channel while weighting the typically more persistent upper channel when retrieving place and yaw hypotheses.

The contributions of this work are threefold:
\begin{itemize}
    \item We introduce a dual-envelope polar descriptor and mask-aware non-uniform matcher that preserve both lower/middle and overhead surfaces while limiting the influence of cross-session changes in the lower layer.
    \item We construct a practical shared map/query front end for this descriptor. Independent gravity canonicalization removes roll/pitch disagreement while retaining free yaw, and a measured descriptor-origin height expresses returns in a common ground-relative frame. A constrained, cell-balanced adaptive split is estimated once from the map and reused by every query, removing per-environment threshold tuning and reducing to single-layer SC when vertical separation is unsupported.
    \item We validate the implemented pipeline through controlled ablations, retrieval comparisons with paired significance tests and precision-oriented acceptance metrics, measured point-cloud cases, and continuous replay across indoor, mixed outdoor/indoor, mounting-height, and outdoor sequences. The implementation, descriptors, query clouds, and evaluation protocols accompany this preprint.
\end{itemize}

%%%%%%%%%%%%%%%%%%%%%%%%%%%%%%%%%%%%%%%%%%%%%%%%%%%%%%%%%%%%%%%%%%%%%%%%%%%%%%%%
\section{Related Work}

\subsection{Global LiDAR Descriptors}

Handcrafted global descriptors remain attractive for onboard LPR because they require neither training data nor a GPU. M2DP summarizes multiple point-cloud projections~\cite{he2016m2dp}. SC uses a polar maximum-height image whose column shifts represent yaw~\cite{kim2018scancontext}; SC++ adds lateral-shift-tolerant variants~\cite{kim2022scancontextpp}. FreSCo moves SC matching to the frequency domain~\cite{fan2022fresco}, LiDAR Iris forms a binary polar signature~\cite{wang2020lidariris}, and SOLiD reorganizes range, azimuth, and elevation for restricted fields of view~\cite{kim2024solid}. Contour Context encodes abstract structural contours~\cite{jiang2023contour}, and RING++ derives a roto-translation-invariant representation with a planar pose estimate~\cite{xu2023ringpp}. None of these prevents an indoor overhead surface from replacing lower structures in the same SC cell; \method\ retains the efficient polar organization while making this vertical competition explicit.

\subsection{Geometric and Cross-Platform Recognition}

Geometric methods use transform-invariant local constellations. STD hashes triangles formed from stable keypoints~\cite{yuan2023std}, and BTC augments triangle geometry with local binary descriptions~\cite{yuan2024btc}. Beyond pure geometry, textual cues read from signage can disambiguate repetitive indoor scenes~\cite{jin2025textual}; \method\ stays geometry-only. Multi-session systems can instead exploit normal-vector consistency during map alignment~\cite{ma2026nvms}; our focus is the retrieval front end before such optimization.

Learned global descriptors include point-set aggregation in PointNetVLAD~\cite{uy2018pointnetvlad}, sparse-voxel features in MinkLoc3Dv2 (MinkLoc-v2)~\cite{komorowski2022minkloc3dv2}, range-image learning in OverlapNet~\cite{chen2020overlapnet}, and the yaw-invariant OverlapTransformer (OT)~\cite{ma2022overlaptransformer}. \method\ is training-free; our mounting-height experiments use the same LiDAR model to isolate vertical geometric mismatch rather than beam-pattern changes.

\begin{figure*}[t]
    \centering
  \subfloat[]{\raisebox{0.0cm}{\includegraphics[width=0.243\textwidth]{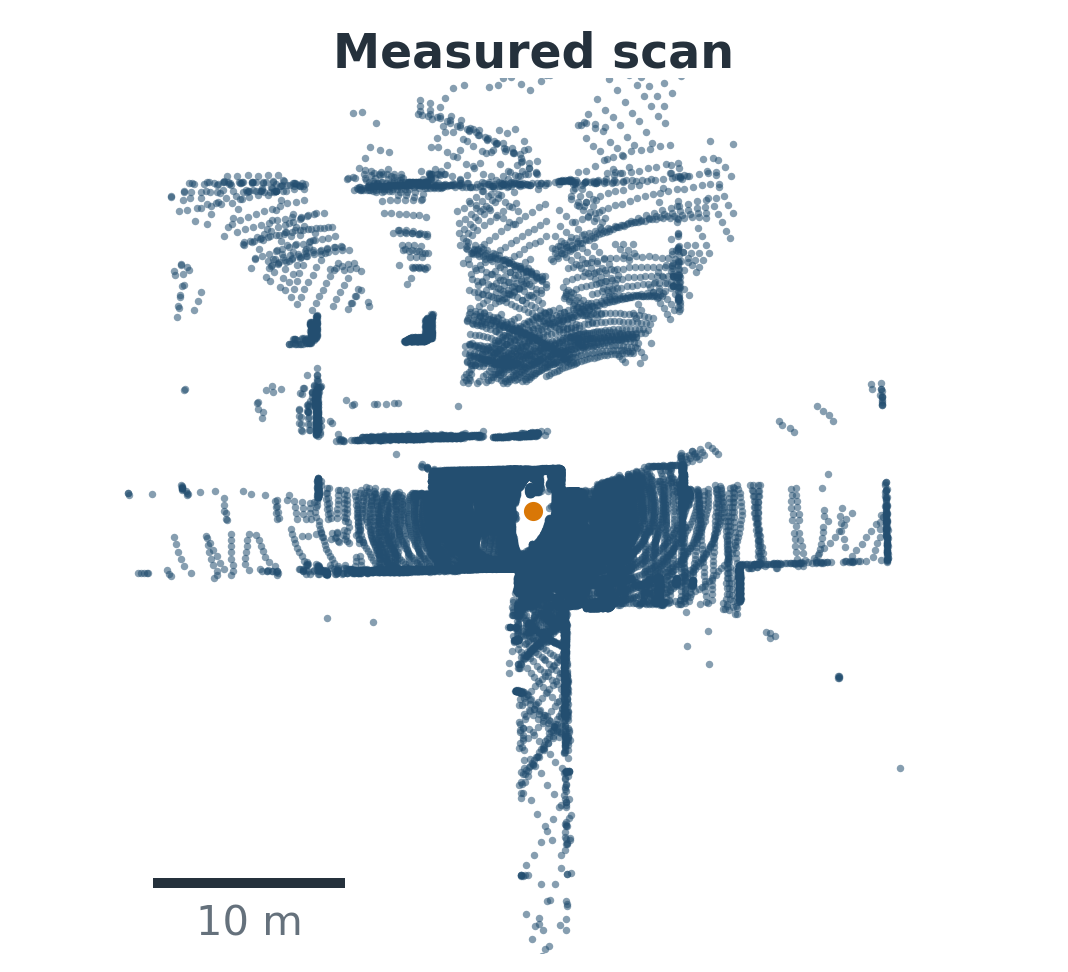}}\label{fig:descriptor_cloud}}
    \hfill
    \subfloat[]{\includegraphics[width=0.195\textwidth]{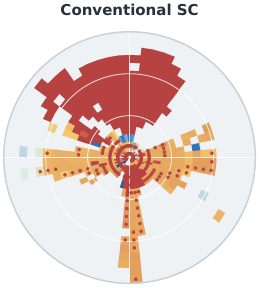}\label{fig:descriptor_sc}}
    \hfill
    \hspace{0.008\textwidth}
    \raisebox{0.18cm}{\textcolor{black!25}{\rule{0.4pt}{3.55cm}}}
    \hspace{0.008\textwidth}
    \hfill
    \subfloat[]{\includegraphics[width=0.490\textwidth]{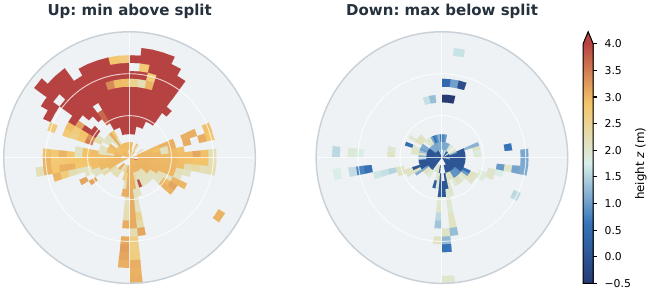}\label{fig:descriptor_updown}}
    \caption{One measured query shown at a common gravity-canonicalized
    reference. (a) White-background top view with a display-only crop of sparse
    range-boundary returns; all valid points still enter the descriptor, and the
    orange marker is its origin. (b) Conventional SC is dominated by overhead maxima in 431 of 467
    occupied cells; 197 of these cells also contain lower/middle returns.
    (c) Up retains the overhead layer and Down separately retains the
    lower/middle layer. Descriptor heights share a color scale
    capped at 4.0~m, with the triangular tip denoting larger values.}
    \label{fig:descriptor}
\end{figure*}

%%%%%%%%%%%%%%%%%%%%%%%%%%%%%%%%%%%%%%%%%%%%%%%%%%%%%%%%%%%%%%%%%%%%%%%%%%%%%%%%
\section{Method}

\subsection{Problem Formulation and Overview}

Let a motion-compensated query frame be
$\mathcal{P}^{q}=\{\mathbf{p}^{q}_{n}\}_{n=1}^{N_q}$, expressed at its reference time in the descriptor-origin body frame. Its inertial estimate supplies a gravity vector $\hat{\mathbf{g}}^{q}$ in that frame. A map database contains keyframe scans $\mathcal{P}^{m}$ and their poses $\mathbf{T}^{w}_{m}\in SE(3)$. For a query, the desired front-end output is a ranked list
\begin{equation}
 \mathcal{Y}^{q}=\left\{\left(m_k,
 \{(d_{k\ell},\Delta\psi_{k\ell})\}_{\ell=1}^{L_k}\right)\right\}_{k=1}^{K_q},
 \quad L_k\leq3,
    \label{eq:output}
\end{equation}
where $m_k$ is a database keyframe and each pair contains a descriptor distance and sector-shift yaw hypothesis; candidates are ranked by $d_k=\min_{\ell}d_{k\ell}$. A geometric verifier then refines the corresponding pose seeds.

Fig.~\ref{fig:method_workflow} summarizes the complete online chain. The LIO
front end deskews one scan and supplies its inertial gravity estimate. Gravity
canonicalization removes roll and pitch while retaining free yaw, and the
dual-envelope descriptor queries a keyframe database for ranked place and yaw
hypotheses. Each forms a gravity-consistent 6-DoF seed for point-to-map ICP.
Only a registration that passes overlap, fitness, and convergence checks
confirms the retrieved place and supplies a geometric constraint to the
loop-closure or global-relocalization back end.
Fig.~\ref{fig:motivation} highlights
the indoor information loss addressed by the dual envelope.

\subsection{Gravity Canonicalization}

We use one LiDAR frame and deskew every return to its reference timestamp using LiDAR-inertial odometry~\cite{xu2022fastlio2}, restoring points transformed to the IMU frame for deskewing back to the descriptor-origin body frame before descriptor construction. Let the measured up direction be $\hat{\mathbf{u}}=-\hat{\mathbf{g}}/\|\hat{\mathbf{g}}\|$. We compute the minimum rotation $\mathbf{R}_{G\leftarrow B}\in SO(3)$ that maps $\hat{\mathbf{u}}$ to $\mathbf{e}_{z}=[0,0,1]^\top$. Because $\mathbf p_n$ is already descriptor-origin centered, its canonical coordinate is
\begin{equation}
    \tilde{\mathbf{p}}_n=\mathbf{R}_{G\leftarrow B}\mathbf{p}_n,
    \label{eq:gravity}
\end{equation}
where $B$ and $G$ denote the body and gravity-canonical frames. This operation fixes the vertical axis without choosing a horizontal heading, so rotations about gravity remain circular shifts of the descriptor sectors, preserving SC's efficient yaw search. Map and query scans are leveled independently, so their roll and pitch need not be known relative to one another.

Gravity canonicalization assumes that the inertial gravity estimate is sufficiently accurate at the scan reference time; sustained acceleration and residual deskew error remain failure modes discussed in Sec.~\ref{sec:discussion}.

\subsection{Map-Adaptive Dual-Envelope Scan Context}

For each canonicalized point $\tilde{\mathbf{p}}=(x,y,z)$, define range $r=\sqrt{x^2+y^2}$ and azimuth $\theta=\operatorname{atan2}(y,x)$. The disk $r\leq r_{\max}$ is divided into $N_r$ rings and $N_s$ sectors. Let $\mathcal{C}_{ij}$ denote the points in ring $i$ and sector $j$. If platform $a$ has descriptor-origin height $h_a$, each point is first expressed by its ground-relative height
\begin{equation}
    z_g=z+h_a.
    \label{eq:physical_split}
\end{equation}
The height $h_a$ is a once-measured platform quantity, not a per-frame fitted ground plane. To avoid tuning the lower/overhead boundary for each environment, we estimate one physical split $\tau_g$ from all map keyframes. Heights are quantized into a histogram in which each polar cell casts at most one vote per occupied height bin, preventing dense surfaces and LiDAR sampling patterns from dominating. Among thresholds in $[\tau_{\min},\tau_{\max}]$ for which both partitions contain at least $\rho_{\min}$ of the votes, we select the constrained Otsu optimum
\begin{equation}
 \tau_g=\arg\max_{\tau}\;
 \omega_{\ell}(\tau)\omega_h(\tau)
 \left[\mu_{\ell}(\tau)-\mu_h(\tau)\right]^2.
 \label{eq:adaptive_split}
\end{equation}
Here $\omega_{\ell,h}$ are the normalized vote fractions and $\mu_{\ell,h}$ their weighted mean heights. Equal-score plateaus use their midpoint, and $\tau_g$ is frozen in the map database so queries reuse the map's definition. Both classification and stored values use $z_g$:
\begin{align}
 E^{\downarrow}_{ij}
 &=\max_{\tilde{\mathbf{p}}\in\mathcal{C}_{ij},\,z_g\leq\tau_g} z_g,
 \label{eq:downenv}\\
 E^{\uparrow}_{ij}
 &=\min_{\tilde{\mathbf{p}}\in\mathcal{C}_{ij},\,z_g>\tau_g} z_g.
 \label{eq:upenv}
\end{align}
Each extremum is used only when its corresponding mask below is one; hence an empty maximization or minimization never enters matching.
Here ``Up'' and ``Down'' name the physical height layers, not the surface-normal directions: Up is the upper/overhead layer and stores its downward-facing minimum, whereas Down is the lower/middle layer and stores its upward-facing maximum. We store one validity bit per cell and channel:
\begin{equation}
\begin{aligned}
M^{\downarrow}_{ij}&=\mathbb{I}\!\left[\exists\,\tilde{\mathbf p}\in\mathcal C_{ij}:z_g\leq\tau_g\right],\\
M^{\uparrow}_{ij}&=\mathbb{I}\!\left[\exists\,\tilde{\mathbf p}\in\mathcal C_{ij}:z_g>\tau_g\right].
\end{aligned}
\label{eq:masks}
\end{equation}
An invalid cell is missing data rather than a zero-height measurement. If the map lacks enough support for two admissible height groups, descriptor construction reduces to single-layer SC. In a mixed trajectory, a jointly absent upper channel is ignored only when at most 5\% of at least three map keyframes within 10~m contain upper observations.

The two envelopes resolve a specific loss of information in maximum-height SC. If a cell contains a shelf edge and a ceiling return, conventional SC retains only the ceiling. Equations~\eqref{eq:downenv}--\eqref{eq:upenv} retain both the shelf's highest visible surface and the ceiling's lowest visible surface, so the Up channel preserves the overhead structure represented by conventional SC while the Down channel prevents it from overwriting lower geometry. In the measured query of Fig.~\ref{fig:descriptor}, overhead returns set 431 of 467 occupied cells, 197 of which also contain lower/middle returns that conventional SC discards.

\begin{figure}[t]
    \centering
    \includegraphics[width=\columnwidth]{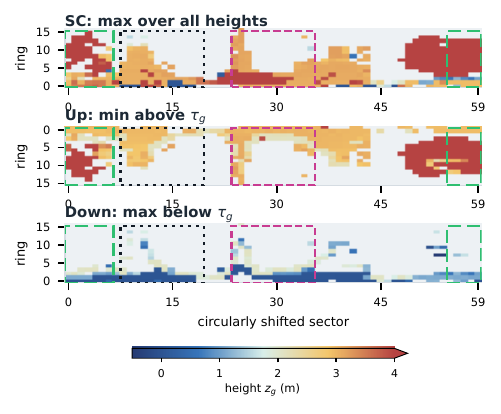}
    \caption{Rectangular SC-style feature maps of the same query as
    Fig.~\ref{fig:descriptor}, sharing one $16\times60$ grid, color scale, and
    circular shift (Up drawn with near rings at the top, as in
    Fig.~\ref{fig:method_workflow}). Up retains SC's overhead support
    (431 cells, 92.3\%); Down restores all 197 lower/middle observations
    hidden by SC's full-height maximum. Data-selected 12-sector windows mark
    three advantages: magenta, the densest mixed-cell region (59 cells);
    green dashed, 49 overhead cells whose stored maximum and visible
    underside differ by a median 2.2~m (kept only by Up); black dotted, lower
    structure spreading 0.90~m where overhead pins SC's spread to 0.35~m. No
    value is interpolated or edited.}
    \label{fig:descriptor_detail}
\end{figure}

Fig.~\ref{fig:descriptor_detail} expands the same scan into rectangular SC-style
feature maps. This separates two effects that are difficult to see in the polar
rendering: taking the minimum above the split preserves broad overhead support
while exposing underside-height variation, and taking the maximum below the
split restores the lower/middle pattern hidden by SC's full-height maximum.

The extrema are taken over the voxel-downsampled scan, whose centroids control point density while preserving a clear max/min envelope interpretation.

\subsection{Non-Uniform Dual-Channel Matching}

The stored descriptor retains valid ground-relative heights, and an explicit mask distinguishes a genuine near-zero height from missing data. Write $\mathbf e^{c,a}_{j}=\mathbf E^{c,a}_{:,j}$ for column $j$ of scan $a\in\{q,m\}$. We define $[\operatorname{shift}_s(\mathbf v)]_j=v_{(j-s)\bmod N_s}$; thus query sector $j$ is compared with map sector $j-s$. Their joint mask is $\mathbf m^c_j(s)=\mathbf M^{c,q}_{:,j}\odot\mathbf M^{c,m}_{:,j-s}$ for $c\in\{\uparrow,\downarrow\}$. A small offset $b$, used only on valid entries in retrieval, conditions near-ground cosine vectors without changing the stored SCD:
\begin{equation}
\bar{\mathbf e}^{c,q}_{j}(s)=\mathbf m^c_j(s)\odot
(\mathbf e^{c,q}_{j}+b\mathbf 1),\;
\bar{\mathbf e}^{c,m}_{j-s}(s)=\mathbf m^c_j(s)\odot
(\mathbf e^{c,m}_{j-s}+b\mathbf 1),
\label{eq:retrieval_offset}
\end{equation}
so only jointly observed rings enter the height cosine and absent cells remain masked. To prevent a small coincident subset from producing a spuriously high cosine, we also measure the normalized overlap of the two binary masks,
\begin{equation}
 \gamma^c_j(s)=
 \frac{\|\mathbf m^c_j(s)\|_0}
 {\sqrt{\|\mathbf M^{c,q}_{:,j}\|_0
              \|\mathbf M^{c,m}_{:,j-s}\|_0}}.
 \label{eq:mask_similarity}
\end{equation}
Equation~\eqref{eq:mask_similarity} is evaluated only for $j\in\mathcal J_c(s)$. The coefficient equals one for identical valid-ring support and decreases when only a subset overlaps. With $n_{\min}=2$, let $\mathcal Q_c=\{j:\|\mathbf M^{c,q}_{:,j}\|_0\geq n_{\min}\}$, $\mathcal C_c(s)=\{j:\|\mathbf M^{c,m}_{:,j-s}\|_0\geq n_{\min}\}$, and $\mathcal J_c(s)=\{j:\|\mathbf m^c_j(s)\|_0\geq n_{\min}\}$. Columns with either cosine norm below a numerical tolerance are also excluded from $\mathcal J_c(s)$. For a comparable channel ($|\mathcal J_c(s)|>0$), the sector-support coefficient
\begin{equation}
 \eta_c(s)=\frac{|\mathcal J_c(s)|}
 {\sqrt{|\mathcal Q_c|\,|\mathcal C_c(s)|}}
 \label{eq:sector_support}
\end{equation}
measures the agreement of the supported-sector masks; its denominator is then nonzero. Because $\gamma^c_j$ already penalizes missing support at the ring level, we apply the square root of $\eta_c$ to avoid counting partial visibility twice while still suppressing candidates supported by only a few accidental sectors. The channel distance is
\begin{equation}
 \delta_c(s)=1-\frac{\sqrt{\eta_c(s)}}{|\mathcal{J}_c(s)|}
 \sum_{j\in\mathcal{J}_c(s)}
 \gamma^c_j(s)
 \frac{\langle\bar{\mathbf e}^{c,q}_{j}(s),
 \bar{\mathbf e}^{c,m}_{j-s}(s)\rangle}
 {\|\bar{\mathbf e}^{c,q}_{j}(s)\|_2
  \|\bar{\mathbf e}^{c,m}_{j-s}(s)\|_2},
 \label{eq:channel_distance}
\end{equation}
where shifts wrap modulo $N_s$. A jointly absent Down channel is omittable; a jointly absent Up channel is omittable only under the locally overhead-sparse rule above. Let $\mathcal A(s)$ contain the comparable channels. If any positive-weight channel is neither comparable nor omittable, we set $d(s)=1$; otherwise
\begin{equation}
 d(s)=
 \frac{\sum_{c\in\mathcal A(s)}w_c\delta_c(s)}
      {\sum_{c\in\mathcal A(s)}w_c},
 \quad (w_{\downarrow},w_{\uparrow})=(w_\ell,w_h).
\label{eq:weighted_distance}
\end{equation}
If $\mathcal A(s)=\varnothing$, the shift is discarded. Thus omitted channels
are removed from both the sum and its weight normalization.

The lower/middle channel is often more discriminative but is also more exposed to people, chairs, carts, and open doors. We therefore select $(w_{\ell},w_h)=(0.3,0.7)$ on the IH+G pilot and freeze it for every other dataset. This bounded robustness choice is not a dynamic-object model, and lower structures remain in the descriptor.

As in SC, a compact ring key shortlists $K$ map entries. It aggregates masked sectors from both channels and remains yaw invariant because the sector index is removed. For each entry, a sector-key comparison over all circular shifts gives a coarse alignment, and the full dual-envelope distance is evaluated only in released SC's local 10\% window: seven shifts for $N_s=60$.

\subsection{Yaw Estimation}

We preserve the conventional SC decision order. Let $\mathbf{k}^{q}$ and $\mathbf{k}^{m}$ denote the query and map sector keys. For each shortlisted map entry, define
\begin{equation}
 \begin{aligned}
 e_k(s)&=\left\|\mathbf{k}^{q}
       -\operatorname{shift}_{s}(\mathbf{k}^{m})\right\|_2,\\
 \bar{s}&=\underset{s\in\{0,\ldots,N_s-1\}}{\arg\min}\ e_k(s),
 &r&=\operatorname{round}(0.05N_s),\\
 \mathcal{S}(\bar{s})&=\{\bar{s}-r,\ldots,\bar{s}+r\}.
 \end{aligned}
 \label{eq:yaw_search}
\end{equation}
where indices wrap modulo $N_s$. We evaluate $d(s)$ for $s\in\mathcal{S}(\bar{s})$ and retain the best three, preferentially separated, yaw hypotheses. For any retained $s^{*}$, the descriptor sector shift is
\begin{equation}
    \widehat{\Delta\psi}=\frac{2\pi s^{*}}{N_s}.
    \label{eq:yaw}
\end{equation}

\subsection{6-DoF Verification and Complexity}

For candidate $m$ in the gravity-aligned map frame $w$, let $\psi_m^G$ be its stored canonical heading and let
$\mathbf{R}_{G\leftarrow B}^{q}$ rotate the query body frame into its
gravity-canonical frame. The descriptor does not infer roll and pitch from
height values; the measured gravity direction constrains them, while
Eq.~\eqref{eq:yaw} estimates the remaining heading offset. Together with the
candidate translation $\mathbf{t}_m^w$, these terms form the complete initialization
\begin{equation}
 \widehat{\mathbf{R}}_{w\leftarrow B}^{q}
 =\mathbf{R}_{z}\!\left(\psi_m^G-\widehat{\Delta\psi}\right)
  \mathbf{R}_{G\leftarrow B}^{q},\qquad
 \widehat{\mathbf{t}}_{w\leftarrow B}^{q}=\mathbf{t}_m^w .
 \label{eq:icp_seed}
\end{equation}
No descriptor-derived vertical correction is applied; ICP thus starts from a
gravity-consistent 6-DoF pose rather than searching attitude from scratch.
Point-to-map ICP then refines all six degrees of
freedom and rejects candidates that fail overlap, fitness, or convergence
checks; the retrieval benchmarks score descriptor rankings before ICP.

Envelope construction is linear in the scan point count. A compact float-and-bitset representation of $2N_rN_s$ values and two masks requires $\approx$7.9~kB per keyframe at $16\times60$; database memory and ring-key shortlisting therefore scale linearly in keyframe count, as in SC. After ring-key retrieval, yaw matching has the conventional SC circular-shift cost. The runtime experiment reports query descriptor construction and database matching/ranking separately from geometric verification.

%%%%%%%%%%%%%%%%%%%%%%%%%%%%%%%%%%%%%%%%%%%%%%%%%%%%%%%%%%%%%%%%%%%%%%%%%%%%%%%%
\section{Evaluation Protocol}

\subsection{Datasets}

The in-house (IH) two-session pilot uses front and rear MID-360 units,
calibrated into a common body frame and fused into the single observation
consumed by FAST-LIO and every descriptor. A loop-corrected mapping bag
supplies 376 database keyframes; a separately recorded localization bag
supplies 322 query keyframes, of which 320 have a database keyframe within
the predeclared 2~m positive radius. Query positions come from an independent
prior-map localization replay (pseudo-reference positions, not survey ground
truth).

We evaluate public cross-sequence transfer on the RTK-SLAM
Construction Hall data~\cite{zhang2026rtkslam}\footnote{Dataset:
\url{https://www.isprs.org/resources/datasets/benchmarks/RTK-SLAM/Default.aspx}}.
Both MID-360 traversals begin and end outdoors and pass through the
GNSS-degraded hall interior (0.48~km and 0.39~km routes); Sequence~1 forms
the map/database and Sequence~2 supplies queries. The shared experimental 2~m sampling below yields 189 map keyframes and 148
single-frame queries. We align each
FAST-LIO trajectory to a metric frame using valid GNSS segments and count a
retrieval as correct when any returned map keyframe is within 5~m of the
query; all 148 queries have map overlap under this rule, and this
trajectory-level protocol complements the dataset's sparse surveyed control
points.

The indoor mounting-height deployment uses one hand-carried run (H1) as the
map and independently recorded hand-carried (H2) and vehicle-mounted (V1)
runs as queries. All use the same MID-360, so the comparison isolates session,
occlusion, attitude, and mounting height from beam-pattern variation.

For public indoor transfer with surveyed per-frame truth, we use the M2DGR
hall sequences~\cite{yin2022m2dgr} (VLP-32C ground robot): hall\_04 forms the
33-keyframe map and cross-day hall\_02 supplies 28 eligible queries. Sessions
align to their Leica MS60 tracks by jointly estimated truth clock offsets and
prism lever arm (4.4/4.9~cm RMSE); the two days' station frames are
registered by wall-point ICP (10.3~cm median); measured origin heights are
0.79/0.80~m.

For outdoor generalization, we use the OS0-128 Quad-Easy traversal from the
Newer College extension~\cite{zhang2021newerextension}: its first temporal
half forms the map, the second half supplies queries, and the same 2~m
sampling produces 59 map keyframes and 60 queries, all with a map keyframe
within the 5~m correctness radius under the published ground truth.

\subsection{Sampling and Metrics}

For all retrieval experiments, a new experimental keyframe is selected after
2~m of 3-D translation (first frame retained); time and yaw triggers are
disabled only for this evaluation. Each query is one deskewed frame (nominally
0.1~s). IH, the indoor deployment, and M2DGR use a 2~m positive radius; CH and
Newer College use 5~m, matching CH's trajectory-level GNSS reference. We report
Recall@1/@5 before geometric verification; failed descriptors and empty
candidate sets count as failures. Because query counts are limited, we quantify
uncertainty with Wilson 95\% intervals for Recall@1 (half-widths
$\approx\pm5$ points on IH, $\pm8$ on CH, and $\pm12$ on the 57--58-query
deployments) and paired McNemar exact tests on shared queries. Sweeping an
acceptance threshold over each query's top-1 confidence yields
precision--recall curves, F1max, and AUPR. Runtime covers query construction
and retrieval/ranking from a preloaded database, excluding I/O, map
construction, and ICP.

%%%%%%%%%%%%%%%%%%%%%%%%%%%%%%%%%%%%%%%%%%%%%%%%%%%%%%%%%%%%%%%%%%%%%%%%%%%%%%%%
\section{Experiments}

\subsection{Baselines and Fairness}

Table~\ref{tab:settings} lists the core descriptor and retrieval settings.
We compare SC~\cite{kim2018scancontext}, SC++~\cite{kim2022scancontextpp},
SOLiD~\cite{kim2024solid}, M2DP~\cite{he2016m2dp}, LiDAR
Iris~\cite{wang2020lidariris}, RING++~\cite{xu2023ringpp}, and
BTC~\cite{yuan2024btc}. OT and MinkLoc-v2 use the official KITTI- and
Oxford-trained checkpoints, respectively, without fine-tuning. Every native
single-frame baseline receives the same deskewed body-frame cloud and spatial
crop; the explicitly labeled ``+G'' diagnostic additionally receives our
gravity canonicalization, isolating the shared front end. BTC retains its
native accumulated-input protocol and is marked separately.
STD~\cite{yuan2023std}, evaluated under the same accumulated protocol,
retrieved near-zero indoor recall (2.2/7.7\% on IH, 0.7/0.7\% on CH), so BTC
represents the keypoint-triangle family in Table~\ref{tab:retrieval}.

\begin{table}[t]
\caption{Core descriptor and retrieval settings.}
\label{tab:settings}
\centering
\scriptsize
\setlength{\tabcolsep}{3.5pt}
\begin{tabular}{@{}p{0.55\columnwidth}p{0.37\columnwidth}@{}}
\toprule
Parameter & Value \\
\midrule
\multicolumn{2}{@{}l}{\textit{Descriptor construction}} \\
Rings / sectors & 16 / 60 \\
Maximum radius & 30 m \\
Voxel size & 0.25 m \\
Adaptive split range / bin & $[1.5,4.5]$ m / 0.1 m \\
Adaptive split: support / map keyframes & 5\% / 20 \\
\addlinespace[1pt]
\multicolumn{2}{@{}l}{\textit{Descriptor retrieval}} \\
Retrieval-only offset $b$ & 0.1 m \\
Minimum joint rings $n_{\min}$ & 2 \\
Sector-support exponent & $1/2$ \\
Down/Up weights $(w_{\ell},w_h)$ & $(0.3,0.7)$ \\
Ring-key shortlist $K$ & 100 \\
Retained yaw hypotheses & 3 \\
\bottomrule
\end{tabular}
\end{table}

\subsection{Cross-Session Place Retrieval}

Table~\ref{tab:retrieval} compares the in-house pilot (IH), the public
mixed outdoor-to-indoor Construction Hall (CH) sequence, and two
mounting-height deployments, with Newer College (NC) as an outdoor transfer
check. IH contains 376 map keyframes and 320 eligible queries; CH contains
189 and 148. IH is already ground-referenced and uses $h=0$, while CH ground
audits give $h=1.3$~m for both traversals. The adaptive map splits are 2.1~m
(IH), 4.0~m (CH), and 2.4~m (H1), and each query reads the split stored in its
map database. BTC uses the current scan and nine causal predecessors
registered into its current frame; excluding incomplete prefixes leaves 319
IH, 139 CH, and 59 NC queries. Bold and underline denote the best and second-best
single-frame values in each column. Latency is measured on an Intel Core
Ultra~5 225H using 92 preloaded queries against the 2,574-keyframe production
IH prior-map database held in memory.

\begin{table*}[t]
\caption{Indoor and outdoor retrieval (R@1/R@5, \%).}
\label{tab:retrieval}
\centering
\scriptsize
\setlength{\tabcolsep}{2.2pt}
\renewcommand{\arraystretch}{1.02}
\begin{tabular}{lccccccccc}
\toprule
& \multicolumn{2}{c}{IH pilot ($n_q=320$)}
& \multicolumn{2}{c}{CH public ($n_q=148$)}
& \multicolumn{2}{c}{Indoor deployment (+G)}
& \multicolumn{1}{c}{M2DGR (+G)}
& \multicolumn{1}{c}{NC outdoor (+G)}
& \multicolumn{1}{c}{Query latency ($n_q=92$)} \\
\cmidrule(lr){2-3}\cmidrule(lr){4-5}\cmidrule(lr){6-7}
\cmidrule(lr){8-8}\cmidrule(lr){9-9}\cmidrule(lr){10-10}
Method & Native & +G & Native & +G
& H1$\rightarrow$H2 ($n_q=58$) & H1$\rightarrow$V1 ($n_q=57$)
& $n_q=28$ & $n_q=60$ & Median (ms) $\downarrow$ \\
\midrule
SC & 59.1/74.1 & 62.5/79.4 & \best{51.4}/66.9 & 60.1/82.4
   & 39.7/67.2 & 47.4/78.9 & \best{96.4}/\underline{96.4} & \underline{98.3}/\best{100.0} & \best{3.6} \\
SC++ (PC) & 53.4/76.9 & 54.7/77.8 & 39.9/68.9 & 62.8/\underline{87.2}
          & 31.0/51.7 & \underline{50.9}/80.7 & 75.0/\best{100.0} & 96.7/\best{100.0} & \underline{8.8} \\
SOLiD & 35.6/57.5 & 34.4/55.6 & 35.8/\best{75.0} & 42.6/69.6
      & 39.7/60.3 & 36.8/61.4 & \underline{92.9}/\best{100.0} & 53.3/88.3 & 13.3 \\
M2DP$^\dagger$ & 47.8/63.7 & 50.0/64.1 & \underline{48.6}/68.2 & 45.9/63.5
                & 6.9/24.1 & 5.3/28.1 & 46.4/85.7 & 93.3/\best{100.0} & 25.5 \\
LiDAR Iris & 58.1/72.2 & 61.9/76.2 & 42.6/67.6 & \best{66.2}/\best{91.2}
            & \underline{44.8}/\underline{69.0} & \best{61.4}/\best{86.0}
            & 85.7/92.9 & 95.0/\best{100.0} & 515.2 \\
RING++$^\star$ & \underline{62.2}/\best{88.8} & \underline{64.1}/\best{87.8}
                & 34.5/56.8 & 41.2/73.0
                & 31.0/67.2 & 26.3/59.6 & 85.7/\best{100.0} & \best{100.0}/\best{100.0} & 4766.3 \\
OT$^\S$ & 13.4/43.4 & 14.1/34.4 & 12.8/37.2 & 26.4/52.7
        & 12.1/22.4 & 17.5/24.6 & 50.0/85.7 & 71.7/\underline{95.0} & 43.2 \\
MinkLoc-v2$^\P$ & 55.3/78.8 & 56.6/79.1 & 32.4/66.9 & 54.7/82.4
                 & 19.0/51.7 & 21.1/33.3 & 60.7/89.3 & 80.0/93.3 & 401.4 \\
BTC (10 scans)$^\ddagger$ & 48.3/66.1 & 52.0/64.3 & 11.5/23.0 & 15.8/30.9
                           & 25.9/46.6 & 19.3/47.4 & -- & 45.8/59.3 & 93.3 \\
\midrule
\method\ (ours) & \best{69.1}/\underline{86.9} & \best{69.4}/\underline{86.9}
                & \best{51.4}/\underline{70.3} & \underline{64.2}/\underline{87.2}
                & \best{48.3}/\best{70.7} & \underline{54.4}/\underline{82.5}
                & \underline{92.9}/\best{100.0} & 96.7/\best{100.0} & 11.6 \\
\bottomrule
\end{tabular}
\end{table*}

SC, SC++ (PC), SOLiD, and M2DP use audited formula-equivalent CPU
implementations (the SOLiD port matches the official descriptor to a maximum
absolute error of $3.8\times10^{-6}$ on a sample scan); LiDAR Iris and BTC use
official C++ cores with dataset adapters; OT uses the official $64\times900$
spherical projection and released PNG-to-uint8 input conversion; MinkLoc-v2
uses its PointNetVLAD centroid/mean-radius normalization, including the global
coordinate-sign convention, and sparse-voxel model; both retrieve by 256-D L2
distance. \method\ calls the production C++ implementation. All
baseline adapters and configurations are included in the release.
$^\dagger$M2DP follows its released MATLAB formula. $^\star$RING++ is a faithful CPU
port (runtime not representative of its CUDA extension). $^\S$OT: KITTI checkpoint; $^\P$MinkLoc-v2: Oxford-only baseline checkpoint; both are zero-shot. $^\ddagger$BTC is an accumulated-input competitor; its
complete causal ten-scan windows use the common crop, incomplete prefixes are
excluded, and the NC cell uses 59 windows. Failures and empty candidate sets
remain in every aggregate; BTC's causal ten-scan windows were not
constructed for M2DGR.

\method\ gives the best IH Recall@1, 5.3 points above gravity-aligned RING++,
while RING++ leads IH Recall@5 by 0.9 points. Under McNemar's exact test on
the shared queries, the first-choice gain over SC+G is significant
($p=0.002$), whereas the margin over RING++ is consistent across both IH
protocols but not individually significant ($p=0.053$ native, $0.15$ +G). On
CH, \method\ ranks second at both cutoffs behind LiDAR Iris. This complementary
behavior suggests that RING++ favors candidate coverage under horizontal
displacement, whereas the dual envelope improves first-choice discrimination.
CH begins outdoors, crosses the hall, and returns outdoors, demonstrating
operation across a mixed transition but not replacing an outdoor-only
benchmark. In the indoor
deployment, measured descriptor-origin heights of 1.6~m (H) and 0.8~m (V)
express both platforms relative to the ground. \method\ returns finite
candidates for all queries, leads H1$\rightarrow$H2 Recall@5, and ranks second
on H1$\rightarrow$V1; at these query counts the orderings among the leading
methods lie within the interval half-widths and are indicative rather than
individually significant. The retained top-1 yaw hypotheses are also accurate
seeds:
on IH+G, over the 222 correctly retrieved queries, the estimated heading
deviates from the localization reference by $1.9^\circ$ at the median and
$6.5^\circ$ at the 95th percentile, consistent with the $6^\circ$ sector
resolution; rare near-$180^\circ$ ambiguities in symmetric corridors are left
to the geometric verifier. The single-hall M2DGR database saturates most
methods; \method\ stays competitive (92.9/100.0, within one query of SC+G)
while transferring to a spinning 32-beam LiDAR with surveyed truth. This
beam-pattern check is absent from the MID-360 experiments. MinkLoc-v2 exceeds
OT on IH/CH and reaches
80.0/93.3 on NC, yet drops to 19.0--21.1 Recall@1 across indoor mounts despite
per-cloud normalization; both learned baselines expose cross-domain and
cross-mount limits without test-set fine-tuning.

\begin{figure}[t]
    \centering
    \subfloat[SC+G]{\includegraphics[width=0.48\columnwidth]{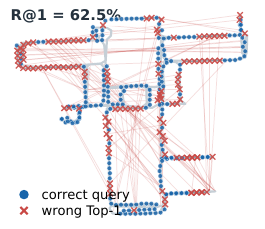}}
    \hfill
    \subfloat[\method+G]{\includegraphics[width=0.48\columnwidth]{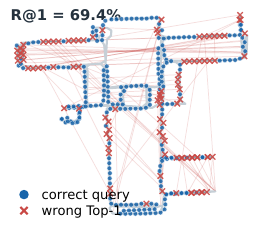}}
    \caption{Top-1 retrievals for all 320 eligible IH queries (shared +G
    protocol): blue circles correct, red crosses failures linked to their
    selected keyframes, gray trajectories.}
    \label{fig:retrieval_distribution}
\end{figure}

Unlike a selected recovery case, Fig.~\ref{fig:retrieval_distribution}
uses all 320 IH queries. The reduction in first-choice failures is distributed
over several repeated branches of the trajectory rather than being
attributable to one visually favorable location.

\begin{figure}[!b]
    \centering
    \includegraphics[width=\columnwidth]{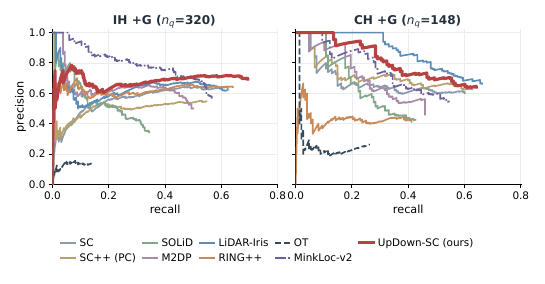}
    \caption{Top-1 acceptance precision--recall under the shared
    gravity-canonicalized protocol. \method\ attains the best F1max/AUPR on
    IH (0.696/0.475, vs.\ 0.643/0.398 for RING++) and the second-best on CH
    (0.644/0.530) behind LiDAR Iris (0.671/0.587) at roughly $40\times$ its
    query latency. BTC lacks a comparable scalar confidence.}
    \label{fig:pr}
\end{figure}

A loop-closure front end must also reject wrong matches, so
Fig.~\ref{fig:pr} sweeps an acceptance threshold over each method's top-1
confidence. \method\ traces the upper envelope on IH and is second behind
LiDAR Iris on CH; without canonicalization it leads AUPR on both IH (0.477
vs.\ 0.370) and CH (0.374 vs.\ 0.354). Full four-condition curves accompany
the release.

\subsection{Outdoor Generalization}

The Newer College column evaluates the gravity-canonicalized front ends on
an outdoor OS0-128 run. \method\ reaches 96.7/100.0\% Recall@1/5,
RING++ 100.0/100.0\%, and SC 98.3/100.0\%: the indoor-oriented dual envelope
remains usable when overhead structure is intermittent, but this 60-query
temporal split is a transfer check rather than evidence of outdoor
superiority. Accumulated-input
BTC reaches 45.8/59.3\% on its 59 complete +G windows (52.5/61.0\% without
+G), showing that gravity canonicalization is not uniformly beneficial to
every geometric descriptor.

\subsection{Continuous Prior-Map Localization}

In a full Sequence~2 replay against the Sequence~1 prior map, the system
first relocalizes in 371~ms and outputs 5,872 trusted poses without a
localization-health warning; the trusted trajectory follows the independently
GNSS-aligned reference with 0.155~m median, 0.308~m 95th-percentile, and
0.349~m maximum position error over the complete traversal. Because these
metrics aggregate the outdoor and indoor portions, they demonstrate operation
across a mixed transition rather than proving outdoor-only superiority.

\subsection{Indoor Mounting-Height Replay}

We additionally map one indoor traversal acquired with the hand-carried
setup and replay two independently recorded query traversals against that
prior: a second hand-carried run and a vehicle-mounted run. The production
prior contains 477 descriptors; the experiment-only 2~m sampling yields 64 map
keyframes and 58/57 overlapping queries with localization-derived
pseudo-reference positions (Table~\ref{tab:retrieval}). Both bags initialize
from one 0.1~s scan, complete without a localization-health warning, and yield
1,609/1,715 trusted poses. These are two deployment cases rather than a
statistically estimated success rate.

\subsection{Ablations and Sensitivity}
\label{sec:ablations}

Table~\ref{tab:retrieval} includes the completed gravity-front-end ablation.
Gravity is not a generic score boost: for \method, it changes IH Recall@1/5
by only $+0.3/0.0$ points, but improves mixed-attitude CH by
$+12.8/+16.9$ points. The varying baseline responses in
Table~\ref{tab:retrieval} show that canonicalization mainly protects
height-structured descriptors when attitude~varies.

\begin{figure}[t]
    \centering
    \subfloat[]{\includegraphics[width=0.72\columnwidth]{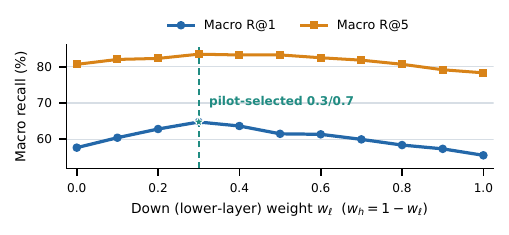}}\\[-0.4em]
    \subfloat[]{\includegraphics[width=0.72\columnwidth]{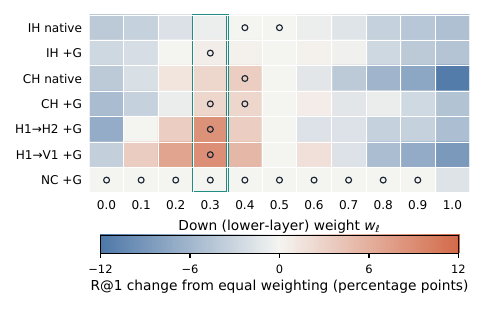}}
    \caption{Channel-weight sensitivity over 11 pairs. (a) Macro Recall@1/5;
    dashed: IH+G-pilot selection $(w_{\ell},w_h)=(0.3,0.7)$. (b) Recall@1
    change from equal weighting for the seven Table~\ref{tab:retrieval}
    conditions; circles mark row maxima and the outline marks the selected pair.}
    \label{fig:weight_ablation}
    \vspace{-0.5em}
\end{figure}

To isolate channel weighting, we reuse the final descriptors and override
only $(w_{\ell},w_h)$, leaving shortlist size, yaw search, masks, and radii
unchanged. Fig.~\ref{fig:weight_ablation} reports the macro-average and all
seven condition-level responses. To avoid test-sequence tuning, the pair is
selected on IH+G pilot Recall@1 and frozen elsewhere. Post hoc, the selected
$(0.3,0.7)$ setting improves macro Recall@1 by 3.3 points over equal weighting,
with its largest gains in the independent indoor replays ($+8.6$ and $+8.8$
points on H1$\rightarrow$H2 and H1$\rightarrow$V1), which are most exposed to
session-dependent occlusion and lower-layer layout change. Both single-channel
extremes are worse, confirming complementarity; the selected pair is a fixed
global compromise, not a uniformly dominant setting.

The adaptive maps select $\tau_g=2.1$, 4.0, and 2.4~m on IH, gravity-aligned
CH, and H1, respectively. Against the fixed 2.5~m ablation, adaptation changes
Recall@1 by at most $+1.0$ point and Recall@5 by $-0.3$ to $+1.8$ points over
the four dual-envelope conditions (e.g., 68.4/87.2 to 69.4/86.9 on IH); its
value is removing per-environment threshold selection rather than a uniform
recall gain. Ground-normalized indoor descriptors provide 100\% candidate
coverage; under a $\pm0.1$~m height-calibration perturbation, H1$\rightarrow$V1
Recall@5 remains 80.7--84.2\%.

\subsection{Runtime and Failure Cases}

Table~\ref{tab:retrieval} reports the common in-memory front-end benchmark:
\method\ requires 11.6~ms at the median and 15.0~ms at the 95th percentile,
and all 92 queries meet a 10~Hz budget, excluding ICP.
Failures arise from repetitive ceilings, glass, multi-level spaces, sparse
overhead sampling, crowds, and excessive horizontal displacement.

%%%%%%%%%%%%%%%%%%%%%%%%%%%%%%%%%%%%%%%%%%%%%%%%%%%%%%%%%%%%%%%%%%%%%%%%%%%%%%%%
\section{Discussion}
\label{sec:discussion}

Two envelopes suit scenes with distinctive but activity-sensitive lower
geometry and stable but repetitive overhead structure. Fixed weights bound
lower-layer influence without removing dynamic objects; excessive overhead
weight may worsen aliasing. Limitations include inertial gravity error, one
measured origin height per platform, a global split missing local modes,
large horizontal displacement, and extrema sensitive to height outliers.
IH uses localization-derived pseudo-reference positions rather than survey
ground truth, while the mounting-height study uses only one LiDAR model;
M2DGR adds a cross-sensor retrieval check, not a mounting study.

%%%%%%%%%%%%%%%%%%%%%%%%%%%%%%%%%%%%%%%%%%%%%%%%%%%%%%%%%%%%%%%%%%%%%%%%%%%%%%%%
\section{Conclusion}

We presented \method, a training-free, gravity-canonicalized dual-envelope
Scan Context. A cell-balanced map distribution determines the split;
complementary envelopes preserve lower/middle and overhead structure; and
masked non-uniform matching retains SC-style retrieval and yaw estimation.
\method\ gives the best first-choice and precision-oriented results on the
in-house pilot, remains competitive on Construction Hall and M2DGR, and
supports continuous prior-map localization. Heterogeneous-sensor invariance
and explicit dynamic-object reasoning remain outside the present claim.

\section*{Acknowledgment}
Generative AI tools (OpenAI Codex, Anthropic Claude Code) assisted with
editing, figures, and tooling; the authors verified all content.

%%%%%%%%%%%%%%%%%%%%%%%%%%%%%%%%%%%%%%%%%%%%%%%%%%%%%%%%%%%%%%%%%%%%%%%%%%%%%%%%
\bibliographystyle{IEEEtran}
\bibliography{references}

\end{document}